%% file: main.tex
\documentclass{article}
\usepackage[utf8]{inputenc}

\usepackage{natbib}
\usepackage{graphicx}
\usepackage{amsmath}
\usepackage{hyperref}
\usepackage{comment}
\usepackage[accepted]{icml2020}
\icmltitlerunning{Trees compensate for misspecification}
\begin{document}

\twocolumn[
    \icmltitle{Decision trees compensate for model misspecification}
    
    \begin{icmlauthorlist}
        \icmlauthor{Hugh Panton$^*$}{in} \,\,\,
        \icmlauthor{Gavin Leech$^*$}{br} \,\,\,
        \icmlauthor{Laurence Aitchison}{br}
    \end{icmlauthorlist}
    
    \icmlaffiliation{in}{\\$^1$Legible Inference, \texttt{explicability.co.uk}}
    \icmlaffiliation{br}{Department of Computer Science, University of Bristol}
    \icmlcorrespondingauthor{Gavin Leech}{g.leech@bristol.ac.uk}
    \icmlkeywords{Machine Learning, interpretability, tree models, linear models}
    \vskip 0.3in
    \date{}
]

% 1. legible is good
% 2. trees not legible
% 3. tree performance can be due to fixable problems
% 4. we have fixed them in a legible model
% 5. huzzah

% .I'm working on ways to make legible models match or outperform the insanely complex models that are rapidly becoming standard anywhere people use math to predict things, and

% .There's an entire list of problems *other* than irreducibly complex behaviour in reality that can incentivize the creation of complex models, and

% .While I'll probably need *some* interactions sprinkled on top of my inherently-interpretable segmented GLMs, I can reduce the demand for model complexity by avoiding/addressing the aforementioned problems, usually through some variation on "if your model assumes the Earth orbits the Sun instead of vice-versa you can get away with using fewer epicycles"."

%\maketitle
\printAffiliationsAndNotice{\icmlEqualContribution }%\icmlEqualContribution} 
\begin{abstract}
\vspace{2mm}
The best-performing models in ML are not interpretable. If we can explain why they outperform, we may be able to replicate these mechanisms and obtain both interpretability and performance.

One example are decision trees and their descendent gradient boosting machines (GBMs). These perform well in the presence of complex interactions, with tree depth governing the order of interactions. However, interactions cannot fully account for the depth of trees found in practice.

We confirm 5 alternative hypotheses about the role of tree depth in performance in the \emph{absence} of true interactions, and present results from experiments on a battery of datasets. Part of the success of tree models is due to their robustness to various forms of mis-specification.

We present two methods for robust generalized linear models (GLMs) addressing the composite and mixed response scenarios.
\end{abstract}

\section{Introduction}

% TODO
%While deep learning has achieved enviable performance, we still need interpretable methods in a variety of practically relevant settings... A common approach to high-performing interpretable methods is GBMs. But high-performing GBMs are often very deep, and this depth compromises interpretability. We find that depth arises in interpretable models such as GBMs because of violated assumptions... We give a series of strategies for mitigating...

Some models are inherently interpretable, for instance sparse linear models e.g.
\begin{align}
\begin{split}
    \mathrm{price} &= \$3000 - \$1000 \times \mathrm{diesel} 
    + \$50 \times \mathrm{red} \\&\quad+ \$100 \times \mathrm{economy} - \$450 \times \mathrm{years}
\end{split}
\end{align}

Why is this interpretable? There are a small number of parameters; the parameters have a natural interpretation on a single scale; the model is \textit{decomposable} (i.e.\! each parameter can be interpreted independently) \citep{lipton2018mythos}; the response is a straightforward function of the parameters.  

However, most models are less straightforward. A typical tree-based model, as produced by the popular library XGBoost \citep{Chen_2016} with default tree-depth 3 and default tree number 100, could be depicted in full:

% TODO vectorise
\begin{figure}[ht!]
\centering
\includegraphics[scale=0.35]{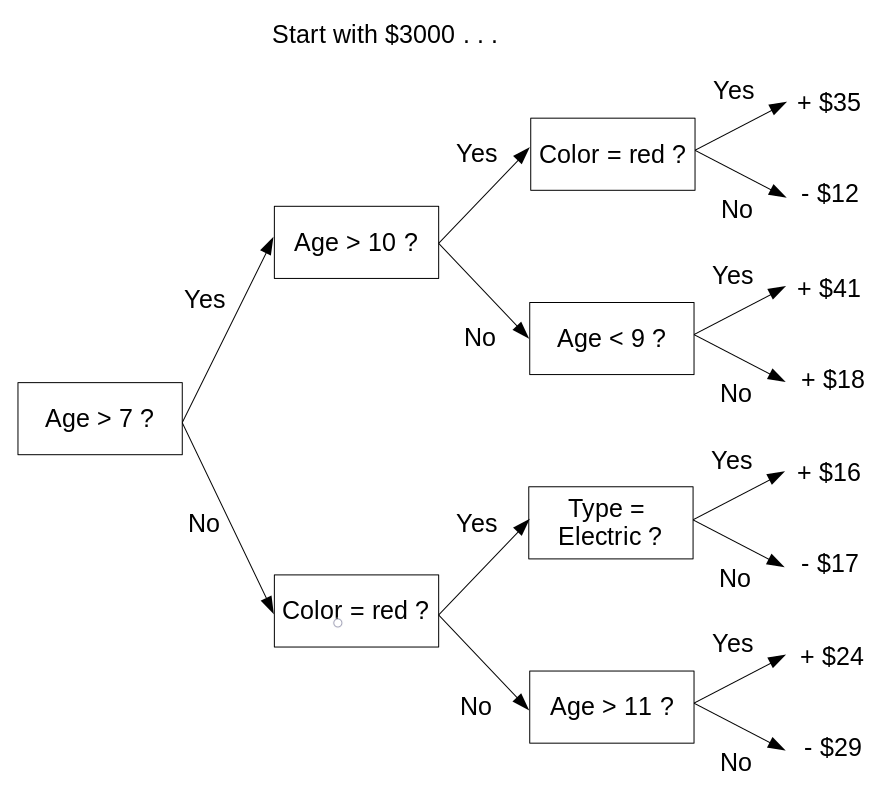}
\includegraphics[scale=0.3]{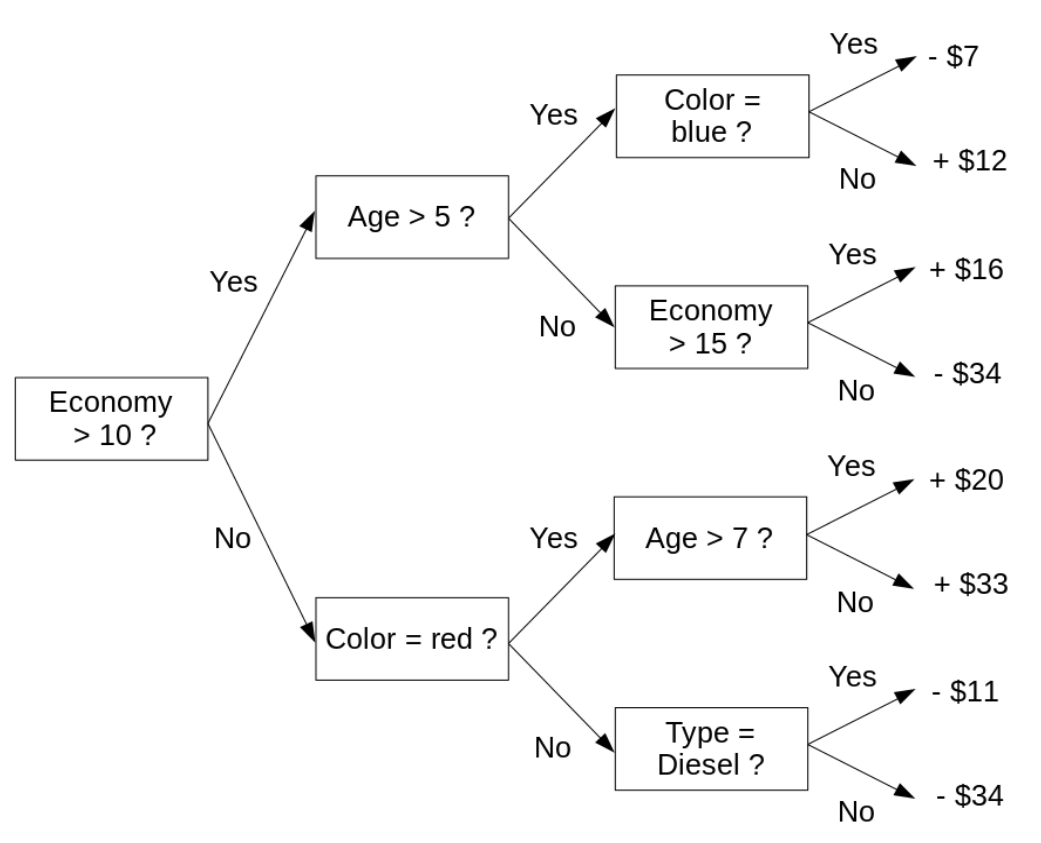}
\caption{\,\,Two of the 100 trees in a typical GBM, depth $d=3$.}
\label{fig:universe}
\end{figure}

but only by also plotting the 98 other trees. This is not interpretable, in several senses. The splits are not decomposable. It is not \textit{simulatable}: no one could understand the global behaviour even given this complete specification. (See \citet{ultrastrong} for an operationalisation of simulatability, as `comprehensibility'.) We are thus forced to use post-hoc model explanation to obtain a (limited) sense of how they work. 

The main obstacle to interpretability is interaction. If the component trees had depth $d\!=\!1$, the ‘decision stumps’ of \citet{IBA1992233}, then partial dependence offers a complete feature-wise summary of the model's behaviour \citep{friedman2001greedy}. For $p$ features used in the splits, an array of $p$ partial dependence plots are unlikely to be as simulatable as our linear equation (1), but it remains possible to see exactly how each feature affects the response.

However, in a complex system, the effects of variables are modified by other variables. Increasingly, datasets are large enough and signals strong enough that we actually can meaningfully model interactions \citep{efron}. In these cases, models without the capacity to model interactions will underperform those which can.

We give five scenarios in which problems besides interaction cause high-depth models to outperform, and show how they could be addressed without using $d>1$. These support the following hypotheses by giving existence proofs:
\begin{enumerate}
    \item These problems may be present where tree-based models are applied, and are at least somewhat responsible for the performance gap between high- and low-tree-depth models.
    \item For a given dataset, the gap between the performance of conventional tree-based models and the \textit{best} interpretable model may be smaller than a standard grid-search suggests.
    \item For a given dataset, the maximum depth required for optimum performance by a tree model may be lower than a standard grid-search suggests.
    \item Models which appear to benefit from $d>1$ might be matched by interpretable models, once these problems are addressed.
\end{enumerate}

The code for the generators, the illustrative GBMs, and our \textit{multiresp} solution is \href{https://osf.io/7hszf/?view_only=678a7ea868d747efbe04ae3e13b4d896}{here}.

\section{Background}

We use GBMs as an exemplar of less interpretable models, and GLMs as an exemplar of interpretable models.

\subsection{Gradient Boosting Machines}

% \subsection{Generalized Linear Models}

% TODO

\section{Related Work}

Machine learning practitioners have found that high-depth tree ensembles perform well; 10-depth ensembles are common, and often win Kaggle contests, thus passing a relatively strict test of generalisation \citep{zhang}. (Note also the unproblematic 11-depth and 21-depth cases in \citet{friedman2001greedy}'s original GBM experiments.) Most accounts of tree success invoke  interaction modelling, or equivalently data segmentation \citep{efron}. This is incongruous when we consider that this claims a ninth-order interaction, and of sufficient power to have estimated this interaction well. There is a classical relationship between interaction effect size, power, and sample size: for a two-way interaction $x_1x_2$ with effect size half that of the main effect, it takes 16 times the sample size to estimate the effect of $x_1x_2$, maintaining precision (\citet{gelman2020regression}, chp 16.4). The standard error compounds as the order of the interaction increases, and as the size of the interaction effect decreases. Some very large datasets and some unusually large interaction effects may pass this bar, but in most cases we should not expect high-order interaction estimates to be justified. So the ubiquity of high-depth tree models, even when datasets and interaction effects are small, presents us with a puzzle.

% maybe add more background for other 4 cases (if above is 4.1)

Little past work has touched on our main topic, the robustness to mis-specification of trees. Linear models can outperform trees in the absence of mis-specification \cite{venkatasubramaniam2017decision}. Some work has noted the ability of tree learning to handle mis-specification \cite{tan2019bayesian}; \citet{kreif2016evaluating}, a medical case study showed that the CART algorithm reduced bias relative to maximum likelihood in the presence of mis-specification.

\section{Methods \& Results}

What \textit{non-interaction} reasons might a tree-based model have for high tree-depth? We give 5 synthetic scenarios and show how tree depth can improve performance. Code for the data generators and GBMs can be found \underline{\href{https://github.com/H-B-P/treedepth}{here}}. Four of the scenarios work with Poisson responses, since three of our hypotheses are shown through multiplicative processes, and Poisson is the simplest model with a canonical log-link. Since our data have known distributions, the results are given without error bars.

\subsection{High tree-depth can compensate for insufficient learning}
\label{scenario:insuff_learn}

Consider a data generating process (\href{https://github.com/H-B-P/treedepth/blob/main/learn_rate/gen.py}{code}):
\begin{align*}
    m &= 10 \\
    n &= 1000 \\
    %\theta_i &\sim \mathrm{Beta}(2, 2) \\
    \theta &= 0.5 \\
    x_i &\sim \mathrm{Bern}(\theta) \quad \mathrm{for }\,\, i=1, ..., m \\
    y_j &\sim \sum^m_{i=0} x_i + \mathcal{N}(0,1)
     \quad \mathrm{for }\,\, j=1, ..., n 
\end{align*}
e.g. flipping ten fair coins a thousand times apiece. The explanatory variables are the results of the ten coin flips; the response variable is the number of heads in each set of ten flips, plus Gaussian noise. 

We model the resulting data using XGBoost with varying $d$; the learning rate and number of trees are specified, but otherwise we use default parameters. Despite there being no interactions between variables, and despite the form of this process being ideal for additive (unity-linked) Gaussians (e.g. decision stumps, $d\!=\!1$), higher-depth models obtain consistently better test scores (Table \ref{tab:scen1_100rounds}).

\begin{table}[!ht]
    \centering
\begin{tabular}{|c|c|c|}
\hline max\_depth $d$ & MAE & RMSE \\
\hline $d=1$ & $1.967$ & $2.382$ \\
\hline $d=2$ & $1.916$ & $2.304$ \\
\hline $d=3$ & $\mathbf{1.886}$ & $\mathbf{2.248}$ \\
\hline
\end{tabular}
    \caption{Example results from Scenario \ref{scenario:insuff_learn}, with 100 trees}    
    \label{tab:scen1_100rounds}
\end{table}

%Consider the `amount of learning' $L(A, H)$ implicit in a given set of hyperparameters $H$ for a given algorithm $A$ [CITE]. Say that $L$ is the quantity which makes a model underfit when too low and overfit when too high. Here, it has a complex but monotonically increasing relationship with $H$ (tree-depth, tree count, and learning rate).

When other hyperparameters result in undertraining, increasing the tree-depth will make the model less underfit, even if there are no interactions to model.

Alternatively, if we are not underfitting, then adding spurious complexity (such as increased depth) to our model will only make things worse. This is elementary; the practical point is that  subtle undertraining along any hyperparameter could inflate the depth of the best model.

For example, trying to find the best tree-depth varying only tree-depth entangles the search for optimal depth with the search for optimal complexity. To actually find the best tree-depth requires a grid-search over depth and other hyperparameters, and moreover a search over a range which actually contains the optimum for each depth.

As a proxy for the degree of underfitting, we use learning rate (LR), where low LR models are assumed to be underfit.

\begin{table}[!ht]
    \centering
\begin{tabular}{|c|c|c|c|}
\hline & $d=1$ & $d=2$ & $d=3$ \\
\hline  $\mathrm{LR}=0.05$ & 1.0948 & 0.9323 & $\mathbf{0.8541}$ \\
\hline $\mathrm{LR}=0.1$ & 0.9053 & 0.8289 & $\mathbf{0.8164}$ \\
\hline $\mathrm{LR}=0.2$ & $\mathbf{0.7994}$ & 0.8124 & 0.8166 \\
\hline $\mathrm{LR}=0.3$ & $\mathbf{0.789}$ & 0.8099 & 0.8255 \\
\hline $\mathrm{LR}=0.4$ & $\mathbf{0.7883}$ & 0.813 & 0.825 \\
\hline $\mathrm{LR}=0.5$ & $\mathbf{0.7884}$ & 0.811 & 0.836 \\
\hline Best & $\mathbf{0.7883}$ & $0.8099$ & $0.8164$ \\
\hline
\end{tabular}
    \caption{MAE for depth and learning rate, under Scenario \ref{scenario:insuff_learn}}
    \label{tab:varying_LR}
\end{table}

There are two relevant asymmetries. The first, shown in Table \ref{tab:varying_LR}, is that undertrained models tend to be worse than overtrained ones. The second is that high-depth models are capable of modelling lower-depth effects, but low-depth models cannot model higher-depth effects, so excessive depth will tend to be less harmful than insufficient depth. 
%(hence XGBoost's default max depth value of 6).

\subsection{High tree-depth can compensate for incorrect linkages}

Consider a multiplicative response with only two true features: the colour and size of gemstones predicts their price. The true relationship is in Table \ref{tab:gemz}.
\begin{table}[ht]
    \centering
\begin{tabular}{|c|c|c|}
\hline & Size $=$ Small & Size $=$ Large \\
\hline Color $=$ Red & $\$ 100$ & $\$ 300$ \\
\hline Color $=$ Blue & $\$ 200$ & $\$ 600$ \\
\hline
\end{tabular}
    \caption{Hypothetical gemstone pricing}    
    \label{tab:gemz}
\end{table}

An additive (unity-linked) model would need tree-depth $>1$ to fully model price. Start with \$100, add \$100 for blueness, add \$200 for largeness, and add a further \$200 for the interaction of blueness and largeness. But our effects are not additive - a multiplicative (log-linked) model would only need tree-depth of 1: blueness multiplies price by 2, largeness multiplies price by 3, and that exhausts the relation.

Real-world datasets outside the physical sciences are generally complex enough that effects from any given feature will have both additive and multiplicative structure (and more structure besides). This implies that models with a conventional approach to linkage will never be perfect. But we can minimise the impact of this  – and hence the tree-depth required – by using the least incorrect linkage.

\subsection{Tree-depth can compensate for multiple responses}
\label{scenario:comp}

The following illustrates when the response is 

Consider a composite multiplicative process (\href{https://github.com/H-B-P/rolecno/blob/main/treedepth_is_too_damn_high/other_reasons/mixes_risks/gen.py}{code}):
\begin{align*}
    m &= 4 \\
    %x_0 &= 1\\
    \theta &= 0.5 \\
    %\theta_i &\sim \mathrm{Beta}(2, 2) \\
    x_i &\sim \mathrm{Bern}(\theta) \quad \mathrm{for }\,\, i=1, ..., m \\\\
    d_A &= I[x_1 = 1] + I[x_2 = 1] + I[x_3 = 0] + I[x_4 = 0]\\
    h_A &= I[x_1 = 0] + I[x_2 = 0] + I[x_3 = 1] + I[x_4 = 1] \\
    \lambda_A &= 2^{d_A} 0.5^{h_A} \\%x_0 \\
    %y_A &\sim \mathrm{Poisson}(\lambda_A)\\
     %\quad \mathrm{for }\,\, j=1, ..., n 
     %
    d_B &= I[x_1 = 1] + I[x_2 = 0] + I[x_3 = 1] + I[x_4 = 0]\\
    h_B &= I[x_1 = 0] + I[x_2 = 1] + I[x_3 = 0] + I[x_4 = 1] \\
    \lambda_B &= 2^{d_B} 0.5^{h_B} \\
    d_C &= I[x_1 = 1] + I[x_2 = 0] + I[x_3 = 0] + I[x_4 = 1]\\
    h_C &= I[x_1 = 0] + I[x_2 = 1] + I[x_3 = 1] + I[x_4 = 0] \\
    \lambda_C &= 2^{d_C} 0.5^{h_C} \\
    %y_B &\sim \mathrm{Poisson}(\lambda_B)\\
     y &\sim \mathrm{Poisson}(\lambda_A) + \mathrm{Poisson}(\lambda_B) + \mathrm{Poisson}(\lambda_C) \\
     X &= (x_1, x_2, x_3, x_4)
\end{align*}

i.e. we flip four coins per row. Start with a value of 1; double it for every heads in the first two coins or tails in the second two, and halve it for every tails in the first two coins or heads in the second two. Use the result as the $\lambda$ for a Poisson generator, and take the output as response A. Repeat the process with the same coin results with different permutations of the doubling and halving indicators to get response B and C. Your total response, the observed $y$ is the sum of these component responses.

This is contrived, but the point holds for any situation where your response variable can be decomposed into multiple processes which deserve their own models, but all you have access to is the sum of outputs from those processes.

For example, say you are predicting staff injuries caused by lions at a zoo; bites and scratches and headbutts should be modelled separately, but you’re only given the total injury count per lion per year. Or an insurance model predicting claims from customer data, where some claims are fraudulent and some are not, but you only know how many claims each customer made in total. 
%Or you’re predicting human height, which can be decomposed into leg height plus torso height plus head height, but you’re only given the head-to-toe measurement.

%So, in other words, any situation where the thing you’re predicting contains more than one type of thing?

If the effects of the explanatory features were additive instead of multiplicative, they would combine simply, and no benefit to increased tree-depth would be observed. But in our log-linked setup, despite no interactions in the response generator, $d\!=\!3$ models consistently beat $d\!=\!1$ models.

\begin{table}[ht]
    \centering
\begin{tabular}{|c|c|c|}
\hline & MAE & RMSE \\
\hline $d=1,100$ trees & $2.9414$ & $4.0419$ \\
\hline $d=1,200$ trees & $2.9413$ & $4.0596$ \\
\hline $d=3,100$ trees & $\mathbf{1.9583}$ & $\mathbf{2.6376}$ \\
\hline
\end{tabular}
    \caption{Example results }    
    \label{tab:multiresp_result}
\end{table}

(Unlike Scenario \ref{scenario:insuff_learn}, a gridsearch is not required to demonstrate our hypothesis: doubling the tree count and getting no significantly better result is sufficient to show that the improvement from increased tree-depth is not just compensation for underfitting.)

In some sense the above process does have interactions - the component features affect each other’s effects on the total output in ways that cannot be captured by an appropriately-linked linear system. In the relevant sense it does not; there are no interactions between the explanatory features in each subresponse. If we ignore the randomness, we have a linear system:

\begin{align*}
    y &= \exp \left( 
        k_{0a} + k_{1a} x_1 + k_{2a}x_2+k_{3a} x_3+k_{4a}x_4 
        \right) \\
        &+ 
        \exp \left( 
        k_{0b} + k_{1b} x_1 + k_{2b}x_2+k_{3b} x_3+k_{4b}x_4 
        \right) \\
        &+
        \exp \left( 
        k_{0c} + k_{1c} x_1 + k_{2c}x_2+k_{3c} x_3+k_{4c}x_4 
        \right) 
\end{align*}
where $y$ is the composite response, $x_i$ are coin flips, and $k_i$ are the coefficients which define the system. Let $S$ be the number of submodels used; here $S=3$. This implies a much simpler algorithm for the above than high-depth trees: we set up a model of that form, apply an appropriate error distribution, and use gradient descent to learn $k_i$. Call this \textit{multiresp}, multiple response modelling.

This toy case yields a model defined by fifteen $k$-values, so that each feature's effect on $\hat{y}$ can still be seen at a glance. In the presence of categorical and quantitative features (as well as binary ones), we could still use partial dependence plots to fully explain each feature's effect.

This is an analogue of mixture modelling, but for composite responses rather than subpopulations in the features. A person fitting two sub-responses to their lion injury data need not know that some are bites and some are scratches; they can just set $S\!=\!2$ and let the algorithm do the rest. Nor do they need to know $S\!=\!2$ apriori: it's just another hyperparameter to optimise.

If we compare the performance of our \textit{multiresp} approach on our toy composite data, we see that the reasoning does hold: high tree-depth and submodels do equally well here.
\begin{table}[!htbp]
    \centering
\begin{tabular}{|c|c|c|}
\hline & MAE & RMSE \\
\hline $d=1,100$ trees & 2.9390 & 4.0973 \\
\hline $d=1,200$ trees & 2.9370 & 4.1085 \\
\hline $d=3,100$ trees & $\mathbf{2.0208}$ & $\mathbf{2.7222}$ \\
\hline Canonical model, $\mathrm{S}=1$ & 2.9370 & 4.1087 \\
\hline Canonical model, $\mathrm{S}=2$ & 2.4667 & 3.4335 \\
\hline Canonical model, $\mathrm{S}=3$ & $\mathbf{2.0207}$ & $\mathbf{2.7223}$ \\
\hline
\end{tabular}
\caption{Example results from tree-based \& canonical models}
    \label{tab:tree_vs_canon}
\end{table}

% TODO: Hugh, run m=8, d<=6 on this
\textit{multiresp} scales well in features $m$; see Appendix \ref{app:scaling} for further results.

\subsection{Tree-depth can compensate for multiple populations}
\label{scenario:mixture}

Consider a mixture of multiplicative processes:
\begin{align*}
    m &= 12 \\
    %x_0 &= 1\\
    %\theta &= 0.5 \\
    %\theta_i &\sim \mathrm{Beta}(2, 2) \\
    x_i &\sim \mathrm{Bern}(0.5) \quad \mathrm{for }\,\, i=1, ..., m \\
    k &\sim \mathrm{Bern}(0.25) \\\\
    d &= x_1 + x_2 + x_3 \\
    d_A &= x_4  + x_5 + x_6   \\
    d_B &= x_7 + x_8  + x_9  \\
    h &= x_{10} + x_{11} + x_{12}  \\
    f_A &= 2^{2k-1} \\
    f_B &= 2 - 1.5k \\
    \lambda &= 2^{d} \, 0.5^{h}\, f_A^{d_A} \, f_B^{d_B} \\
    %y_B &\sim \mathrm{Poisson}(\lambda_B)\\
    % 2^{2k-1} = 0.5 for k = 0
    % 2^{2k-1} = 2 for k = 1
     y &\sim \mathrm{Poisson}(\lambda) 
\end{align*}

That is, for each row, roll a four-sided die ($k$) and flip twelve coins $\{x_i\}$. Rolling a 1 makes it an A-row, and otherwise it’s a B-row: don’t keep that information in the explanatory variables, but do keep all the flips. The value for $\lambda$ starts at 1; each heads on flips 1-3 doubles $\lambda$ for the row; each heads on flips 4-6 doubles $\lambda$ for an A-row but halves it for a B-row; each heads on flips 7-9 halves it for an A-row but doubles it for a B-row; and each heads on flips 10-12 halves it for any row. Once the halvings and doublings are all applied, we generate our response $y$ with a Poisson generator. 

Despite no interactions between the features provided, the dataset generated will consistently yield better scores with $d=3$ than $d=1$.

\begin{table}[!ht]
    \centering
\begin{tabular}{|c|c|c|}
\hline & MAE & RMSE \\
\hline $d=1,1000$ trees & $1.1661$ & $1.7386$ \\
\hline $d=1,2000$ trees & $1.1661$ & $1.7386$ \\
\hline $d=3,1000$ trees & $\mathbf{1.1425}$ & $\mathbf{1.6981}$ \\
\hline
\end{tabular}
\caption{Example results for Scenario \ref{scenario:mixture}}
    \label{tab:multipop}
\end{table}

Scenario \ref{scenario:comp} addressed how multiple effects could be present in the same row; the above addresses how multiple effects could be present across different rows. This applies to any situation where the datapoints being modelled can’t be approximated as homogenous. 

Returning to our examples: let’s say we predict injuries caused by exotic cats at a zoo, and some are lions and some tigers, and the species differences are profound enough to merit different models, but none of your explanatory variables indicate the species. Or, less fancifully, you’re building an insurance model predicting claims from customer data, and some of your customers are honest while others are frauds, but you have no label for this class. 

For this scenario, the system on its own terms looks like this (with two parameter sets, one each for mixture component $a$ and $b$):
\[ 
    y = \exp \left(k_{0a}+k_{1a}
    x_1+ k_{2a} x_2 + 
    \cdots + k_{12a} x_{12} \right) 
\]
    with probability $f$, and
\[ 
    y = \exp \left(k_{0b}+k_{1b}
    x_1+ k_{2b} x_2 + 
    \cdots + k_{12b} x_{12} \right) 
\]
with probability $(1-f)$

So we can use gradient descent to find the best $k$-values, and end up with only twice the complexity of a normal linear model. Call this a \textit{mixed GLM}.

\begin{table}[!ht]
    \centering

\begin{tabular}{|c|c|c|}
\hline & MAE & RMSE \\
\hline Canonical model, $S=2, \mathrm{f}=0.05$ & 1.3554 & 2.5590 \\
\hline Canonical model, $S=2, \mathrm{f}=0.15$ & 1.3789 & 2.4893 \\
\hline Canonical model, $S=2, \mathrm{f}=0.25$ & $\mathbf{1.4160}$ & $\mathbf{2.4615}$ \\
\hline Canonical model, $S=2, \mathrm{f}=0.35$ & 1.5154 & 2.6799 \\
\hline Canonical model, $S=2, \mathrm{f}=0.45$ & 1.5386 & 2.7053 \\
\hline $d=1,1000$ trees & 1.4628 & 2.6322 \\
\hline $d=3,1000$ trees & $\mathbf{1.4181}$ & $\mathbf{2 . 4 6 5 7}$ \\
\hline
\end{tabular}

\caption{Example results from tree-based and canonical models for the Scenario \ref{scenario:mixture} generator. True $f= 0.25$.}
    \label{tab:multipop_plus_f}
\end{table}

We find that $f$ is best treated as a hyperparameter rather than a parameter. To fit ten subpopulations, $S=10$, we would need nine degrees of freedom, i.e.\! nine $f$s to search. However, increasing the submodels as high as 10 would be counterproductive from the perspective of interpretability; making users look at ten numbers per feature (or, for quantitative and categorical analogues, ten graphs per feature) would only be a slight improvement over making them scan hundreds of trees. If you find yourself modelling ten truly distinct subpopulations, then there may be no inherently interpretable model.

Note that mixed GLMs cannot model the specific case of $f=0.5$ and $\mu_a = \mu_b$.

\subsection{Tree-depth can compensate for missing features}
\label{scenario:missing_sc}

This scenario addresses the behaviour of an idealised log-linked dataset where some of the explanatory variables are absent, but are correlated with those present.

Consider a process:
\begin{align*}
    %n &= 10,\!000\\
    m &= 6 \\
    h_i &\sim \mathrm{Bern}(1/2) \quad \mathrm{for }\,\, i=1, ..., m \\ 
    d_i &\sim \mathrm{Bern}(1/10) \quad \mathrm{for }\,\, i=1, ..., m  \\
    %d &= \sum^m_{i=1} d_i \\
    %h^\prime &= (1 - I[d > 0]) x \\
    %
    %
    a_i &= I[d_i \neq 1]h_i + I[d_i = 1](1 - h_i)   \quad \mathrm{for }\,\, i=1, ..., m \\ 
    \lambda &= 2^{\sum^m a_i} \\
     y &\sim \mathrm{Poisson}(\lambda) \\
     \mathbf{x} &= \{ a_i \} \quad \mathrm{for }\,\, i=1,2,3 
\end{align*}

i.e. For each row, toss six coins and roll six ten-sided dice. Record the opposite of the coin flip on rolling a 1, and record the true result on any other result. For your dependent variable $y$, raise 2 to the power of the number of heads you recorded, then feed the result into a Poisson random number generator as in the last two scenarios. For your explanatory variables, take only the first \textit{three} recorded coin flips. When using these explanatory variables to predict this dependent variable with an appropriately-linked and distributed XGB model, $d=3$ outperforms $d=1$.

\begin{table}[!ht]
    \centering

\begin{tabular}{|c|c|c|}
\hline & MAE & RMSE \\
\hline $d=1,100$ trees & $7.4127$ & $11.4690$ \\
\hline $d=1,200$ trees & $7.4125$ & $11.4692$ \\
\hline $d=3,100$ trees & $\mathbf{7.0407}$ & $\mathbf{11.0973}$ \\
\hline
\end{tabular}

\caption{Example results under missing features.}
    \label{tab:missing_col_results}
\end{table}

A natural idea for bringing this effect into interpretable models is to use observed features to predict the posterior distribution of latent feature values, and then use all features together to predict the response. This is impractical; the problem of detecting missing features in general is clearly ill-defined. But even even if it was implementable, the resulting models would be difficult to interpret: it would require graphs to deduce missing features from present features, and others using absent and present features to predict the response. This would obscure the effect of each individual feature on the final prediction, and so for our purposes it is not desirable.

The practical upshot is instead to emphasise investigating extra features when modelling a log-linked system. Besides that, the effect is not obvious: that using more features – even features highly correlated with those present – can make models less complex is  counter-intuitive.

\subsection{Validating on real data}

To illustrate the competitiveness of \textit{multiresp} and mixed GLMs, we evaluate on . 

% TODO: error bars, bootstrap

\section{Conclusions}

The above illustrate scenarios in which $d>1$ outperforms decision stumps, even when there are no real interactions in the data generator. We find that the learner increases depth to compensate for violated assumptions:

\begin{itemize}
    \item “High tree-depth can compensate for insufficient learning”: if you assume the model needs much less training than it does, higher tree-depth will help.
    \item “High tree-depth can compensate for incorrect linkages”: if you assume the system has a linkage it doesn’t, higher tree-depth will help.
    \item “High tree-depth can compensate for multiple responses”: if you assume single responses in a log-linked system where response is composite, higher tree-depth will help.
    \item “High tree-depth can compensate for multiple populations”: if you assume homogeneity in a heterogeneous population, higher tree-depth will help.
    \item “High tree-depth can compensate for absent explanatory columns”: if you assume you have all data needed to explain a log-linked system when you don’t, higher tree-depth will help.
\end{itemize}

We note that all but the first of these are unsurprising, consequences of the simplification inherent to modelling (outside mechanistically grounded domains):

\begin{itemize}
    \item “High tree-depth can compensate for incorrect linkages”: All linkages are approximations.
    \item “High tree-depth can compensate for multiple responses”: All responses are arbitrarily decomposable.
    \item “High tree-depth can compensate for multiple populations”: All models elide relevant heterogeneity.
    \item “High tree-depth can compensate for absent explanatory columns”: All models have absent explanatory columns.
\end{itemize}

This implies that, where these mis-specifications occur, there is scope to fix them and thereby obtain less-complex interpretable models (e.g. a stump ensemble or a GLM) that performs comparably to uninterpretable deep trees. Note though that merely reducing tree depth may not improve matters much: there's a discontinuity at $d=1$ to $d=2$, at which point partial dependence plots stop being a complete summary of model behaviour.

It is a good thing that a learning algorithm succeeds even when the model is not correct, since misspecification is the norm in social science or industrial contexts. However, when interpretability is a priority (as it arguably should be for models that serve a mass market or policy decisions), our results imply that there is an alternative to automatic compensation: attempt to fix the mis-specification and return to interpretable models.

\bibliographystyle{plainnat}
\bibliography{references}

\input{appendix}

\end{document}

%% file: appendix.tex
\newpage
\onecolumn

\appendix

{\Large{\textbf{Appendices}}}

\section{Trees in the limit}

When all explanatory variables are binary, a saturated XGB model with max\_depth = $d$ approaches an unpenalized generalized linear model with every d-way interaction.

i.e. if $d=2$ and there are three explanatory features, the model $y= \mathrm{XGB}(x_1,x_2,x_3)$
will give the same results as 
\[
    y = g^{-1}
    \left(c 
    + k_1 x_1
    + k_2 x_2
    + k_3 x_3
    + (k_{1,2} x_1 x_2)
    + (k_{1,3} x_1 x_3) 
    + (k_{2,3} x_2 x_3)
    \right)
\]
with inverse link function $g^{-1}$.

This assertion was tested by running an XGB model with depth $d$ on a 2x2x2x2 hypercube grid containing the first 16 prime numbers, and comparing the predictions of that grid to those made by a linear model with every $d$-way interaction. For all $d<4$, the outputs were identical.

% https://github.com/H-B-P/rolecno/blob/main/treedepth_is_too_damn_high/other_reasons/idealized_interaction/linear_model.py
% https://github.com/H-B-P/rolecno/blob/main/treedepth_is_too_damn_high/other_reasons/idealized_interaction/model.py
% https://github.com/H-B-P/rolecno/blob/main/treedepth_is_too_damn_high/other_reasons/idealized_interaction/gen.py

% Apply Everywhere

Every XGB model constructed in Section 3.3, Section 3.4, Section 3.5 and Appendix C had an unpenalized generalized linear model of the same depth created on the same data. In every case, the performance of the linear model was very close to that of the XGB model. Crucially, the key patterns were retained: the performance of the d=3 linear model exceeded that of the d=1 linear model, and the performance of the canonical model matched (for Section 3.3) or outperformed (for Section 3.4 and Appendix C) the best linear model.

This argues against the possibility that outperformance of XGB models by canonical models was due to specifying the wrong hyperparameters for training XGB: there is only one optimal unpenalized generalized linear model with every d-way interaction for a given dataset and value of d.

\subsection{Theoretical bounds on the utility of multiresp models and mixture GLMs}

Consider the dataset used in Section 3.3: one response variable, four explanatory binary variables, generation process optimal for using multiresp with $S=3$.

A $d=4$ linear model would need $2^4=16$ k-values to fully describe how the responses predict y. However, the model created using multiresp only has $3(4+1)=15$ k-values. Having one fewer degree of freedom makes it marginally easier to interpret, and marginally less likely to overfit. (The d=3 linear model actually used - optimal for this scenario because one column affects all responses the same way - can be described with only $2^4-4C4=15$ k-values, so there's no improvement.)

Scaling quickly makes the benefit of canonical models more obvious. A d=8 linear model acting on a dataset with eight binary explanatory features (see Appendix C) would need $2^8=256$ k-values to fully describe how the responses predict y; multiresp with S=3 would only need $3(8+1)=27$ k-values.

% (Note 1: This is almost exactly as close to useless as a thing can be. But that’s because I picked this setup because I wanted the simplest case where td=3 reliably beat td=1.)

% (Note 2: I here assume that mo’ DOF = mo’ problems and that all DOF are created equal. Let me know if either of those assumptions are problematic.)

% N = S = max number of parallel models (that could possibly help)
% conditional on all models being slightly different for all features

Formalizing the above: an ideally multiresp-friendly dataset with $B$ binary explanatory features (or, equivalently, explanatory variables conveying B bits of information), in which no columns affect all sub-responses the same way, can be usefully modelled with a multiresp model with S submodels when and only when the resulting model would have fewer k-values than a $d=B$ linear model.

k-values in maximum max\_depth linear model $= 2^B$

k-values in multiresp model = $S(B+1)$

So when you have B binary features, it’s possible for multiresp to lower the number of k-values when
\begin{align}
    S(B+1) < 2^B
\end{align}
For various values of B, the maximum value of S for a multiresp model is presented in Table \ref{tab:multiresp_N}.

\begin{table}[!htbp]
\centering
\begin{tabular}{|l|l|}
\hline
\multicolumn{1}{|l|}{\textbf{B}} & \textbf{Max useful S} \\ \hline
\multicolumn{1}{|l|}{1}          & 0                     \\ \hline
\multicolumn{1}{|l|}{2}          & 1                     \\ \hline
\multicolumn{1}{|l|}{3}          & 1                     \\ \hline
\multicolumn{1}{|l|}{4}          & 3                     \\ \hline
\multicolumn{1}{|l|}{5}          & 5                     \\ \hline
6                                & 9                     \\ \hline
7                                & 15                    \\ \hline
8                                & 28                    \\ \hline
9                                & 51                    \\ \hline
10                               & 93                    \\ \hline
\end{tabular}
\caption{Relationship of number of covariates and maximum $S$ for multiresp models}
\label{tab:multiresp_N}
\end{table}

% Frankenmodel: mixture model (different intercepts) plus prevalences f and 1-f

For the mixture GLMs from section 3.4, the calculation is slightly different, because the $f$s are also a degree of freedom.

Let $k_F$ be the combined number of k-values and f-values in a mixture GLM:
\begin{align}
k_F = S(B+1) + (S-1) = S(B+2) -1
\end{align}

Mixture GLMs can therefore be useful when
\begin{align}
    S(B+2) -1 < 2^B
\end{align}

For various values of $B$, the maximum value of S for a mixture GLM is Table \ref{tab:frankenmodel_N}.

\begin{table}[!htbp]
\centering
\begin{tabular}{|l|l|}
\hline
\multicolumn{1}{|l|}{\textbf{B}} & \textbf{Max useful S} \\ \hline
\multicolumn{1}{|l|}{1}          & 0                     \\ \hline
\multicolumn{1}{|l|}{2}          & 1                     \\ \hline
\multicolumn{1}{|l|}{3}          & 1                     \\ \hline
\multicolumn{1}{|l|}{4}          & 2                     \\ \hline
\multicolumn{1}{|l|}{5}          & 4                     \\ \hline
6                                & 8                     \\ \hline
7                                & 14                    \\ \hline
8                                & 25                    \\ \hline
9                                & 46                    \\ \hline
10                               & 85                    \\ \hline
\end{tabular}
\caption{Relationship of number of covariates and maximum $N$ for mixture GLMs}
\label{tab:frankenmodel_N}
\end{table}

In practice, we would be unlikely to want to use an $S>3$ multiresp model or mixture GLMs, and the typical dataset used in an industry modelling task has hundreds of explanatory columns conveying millions of bits of information.

In other words: outside a small number of pathological cases unlikely to appear in the real world, multiresp models and mixture GLMs are capable of reducing the degrees of freedom used to model compared to a linear model, and hence to an XGB model.

\section{Deriving the GLM objectives}
 
\subsection{Simple (one-population, basic response) case}

% Vanilla version

PMF for Poisson is

%(1a) 
\begin{equation}
   P = \frac{\lambda^k e^{-\lambda}}{k!} 
\end{equation}

Per-row contribution to deviance is log of PMF, so:
%(1b)
\begin{equation}
D = \ln(\lambda^k e^{-\lambda}) - \ln(k!) = k \ln(\lambda) - \lambda - \ln(k!)
\end{equation}

Differentiate with respect to $\lambda$
%(1c) 
\begin{equation}
    \frac{\mathrm{d}D}{\mathrm{d}\lambda} =         
    \frac{k}{\lambda} - 1 =  (k - \lambda)/\lambda
\end{equation}

Define $\lambda$ by its components in a log-linked model

%(1d) 
\begin{equation}
    \lambda = \exp \left(c + m_1x_1 + . . . + m_Nx_N \right)
\end{equation}
Differentiate with respect to components
%(1e) 
\begin{equation}
    \frac{\mathrm{d}\lambda}{\mathrm{d}c} =         
    \exp \left(c + m_1x_1 + . . . + m_Nx_N \right) = \lambda
\end{equation}
%(1f) 
\begin{align}
    \frac{\mathrm{d}\lambda}{\mathrm{d}m_n} =         
    &= x_n \exp \left(c + m_1x_1 + . . . + m_Nx_N \right)  \\
    &= \lambda x_n
\end{align}
So
%(1g) 
\begin{equation}
    \frac{\mathrm{d}D}{\mathrm{d}c} =         
    \frac{\mathrm{d}D}{\mathrm{d}\lambda}         
    \frac{\mathrm{d}\lambda}{\mathrm{d}c}
    = \frac{k-\lambda}{\lambda} \times \lambda = (k-\lambda)
\end{equation}
%(1h) 
\begin{equation}
    \frac{\mathrm{d}D}{\mathrm{d}m_n} =     
    \frac{\mathrm{d}D}{\mathrm{d}\lambda} 
    \frac{\mathrm{d}\lambda}{\mathrm{d}m_n} =         
    \frac{k-\lambda}{\lambda} \times \lambda x_n = (k-\lambda)x_n
\end{equation}

So feed this into gradient descent.

% (hang on, is this shocking neatness at the end what makes a linkage/distribution pair canonical?)

\subsection{Composite response case}

The sum of two Poisson distributions is another Poisson, so the following largely mirrors the above.

Assume two submodels (extending to $n$ is trivial).
%
%(2a) 
\begin{equation}
   \lambda = \lambda_A + \lambda_B
\end{equation}
Differentiate and apply chain rule:
%
%(2b) 
\begin{equation}
    \frac{\mathrm{d}\lambda}{\mathrm{d}\lambda_A} =     
    \frac{\mathrm{d}\lambda}{\mathrm{d}\lambda_B} =     
    1
\end{equation}

Define $\lambda_A$ (and by symmetry $\lambda_B$) by its components in a log-linked model
%
%(2c) 
\begin{equation}
    \lambda_A = \exp \left(
        c_A + m_{1A}x_1 + . . . + m_{NA}x_N
    \right)
    \label{eq:def_lambdaA}
\end{equation}
Differentiate wrt components
%
%(2d) 
\begin{equation}
    \frac{\mathrm{d}\lambda_A}{\mathrm{d}c_A} =         
    \exp \left(c + m_{1A}x_1 + . . . + m_{NA}x_N \right) 
    = \lambda_A
    \label{eq:diff_lambda_c}
\end{equation}
%(2e) 
\begin{equation}
    \frac{\mathrm{d}\lambda_A}{\mathrm{d} m_{nA}} =     
    x_n \exp(c + m_1x_1 + . . . + m_{NA}x_N) = \lambda_A x_n
    \label{eq:diff_lambda_m}
\end{equation}
So
%
%(2f) 
\begin{equation}
    \frac{\mathrm{d}D}{\mathrm{d}c_A} =         
        \frac{\mathrm{d}D}{\mathrm{d}\lambda} \frac{\mathrm{d}\lambda}{\mathrm{d}\lambda_A}
        \frac{\mathrm{d}\lambda_A}{\mathrm{d}c_A}
        = \frac{k-\lambda}{\lambda} \lambda_A
\end{equation}
%(2g) 
\begin{equation}
    \frac{\mathrm{d}D}{\mathrm{d} m_{nA}} =     
        \frac{\mathrm{d}D}{\mathrm{d} \lambda}
        \frac{\mathrm{d}\lambda}{\mathrm{d} \lambda_A}
        \frac{\mathrm{d}\lambda_A}{\mathrm{d} m_{nA}}
    = \frac{k-\lambda}{\lambda} \lambda_A x_n
\end{equation}

\subsection{Mixture of populations case}

%Gotta start from scratch for this one.

PMF for two Poisson processes, one with prevalence $f$ and one with prevalence $(1-f)$, is
%(3a) 
\begin{equation}
    P = f \times \frac{
    \lambda_A^k \exp(-\lambda_A)}
    {k!}
    + (1-f) \times \frac{
    \lambda_B^k \exp(-\lambda_B)}
    {k!}
\end{equation}
Per-row contribution to deviance is log of PMF, so:
%(3b) 
\begin{equation}
    D = \ln \left(
        f \lambda_A^k \exp(-\lambda_A) 
        + (1-f) \lambda_B^k \exp(-\lambda_B)
        \right) - \ln(k!)
\end{equation}
Differentiate with respect to $\lambda_A$
%(3c) 
\begin{align}
    \frac{\mathrm{d}D}{\mathrm{d}\lambda_A} &= 
    \frac{
        f \times \left(
            k\lambda_A^{k-1} 
            \exp \left(-\lambda_A \right) 
            - \lambda_A^{k-1} \exp(-\lambda_A)
            \right) 
        }
        {
        f \times \lambda_A^k \exp(-\lambda_A) + (1-f) \times  \lambda_B^k \exp(-\lambda_B)
        } \\\\
    &= \frac{
            f \times (k-\lambda_A)\lambda_A^{k-1} \exp(-\lambda_A) 
        }
        { 
            f \times \lambda_A^k \exp(-\lambda_A) 
            + (1-f) \times \lambda_B^k \exp(-\lambda_B)
        }
    \label{eq:diff_D_lambda}
\end{align}
Equations (\ref{eq:def_lambdaA}), (\ref{eq:diff_lambda_c}) and (\ref{eq:diff_lambda_m}) still apply, so all that remains is to chain them to (\ref{eq:diff_D_lambda}):
%Just multiply (3c) by (2d) and (2e) and that’s it.
%
\begin{align}
    \frac{\mathrm{d}D}{\mathrm{d} m_{nA}}
        = \frac{\mathrm{d}D}{\mathrm{d}\lambda_A}
        \frac{\mathrm{d}\lambda_A}{\mathrm{d} m_{nA}}
        = x_n \times
        \frac{
            f \times (k-\lambda_A)\lambda_A^{k} \exp(-\lambda_A) 
        }
        { 
            f \times \lambda_A^k \exp(-\lambda_A) 
            + (1-f) \times \lambda_B^k \exp(-\lambda_B)
        }
\end{align}
\begin{align}
    \frac{\mathrm{d}D}{\mathrm{d} c_{A}} = 
        \frac{\mathrm{d}D}{\mathrm{d}\lambda_A}
        \frac{\mathrm{d}\lambda_A}{\mathrm{d}c_A}
    = \frac{
            f \times (k-\lambda_A)\lambda_A^{k} \exp(-\lambda_A) 
        }
        { 
            f \times \lambda_A^k \exp(-\lambda_A) 
            + (1-f) \times \lambda_B^k \exp(-\lambda_B)
        }
\end{align}

\section{Results from scaling \textit{multiresp}}
\label{app:scaling}

The \textit{multiresp} model in section 3.3 only matches the performance of an XGB model with depth 3, and does not exceed it. This is because we tried to use the simplest synthetic dataset for which an XGB model with depth 3 would consistently outperform an XGB model with depth 1. A slightly more complex dataset, generated as in section 3.3 but with twice as many explanatory variables, was used to compare multiresp and XGB.

\begin{table}[!ht]
    \centering
\begin{tabular}{|c|c|c|}
\hline & MAE & RMSE \\
\hline $d=1,500$ trees & $11.6716$ & $25.8049$ \\
\hline $d=1,1000$ trees & $11.6717$ & $25.8051$ \\
\hline $d=2,500$ trees & $4.1651$ & $7.7415$ \\
\hline $d=3,500$ trees & $3.1354$ & $4.9433$ \\
\hline $d=4,500$ trees & $2.8113$ & $4.2739$ \\
\hline $d=5,500$ trees & $2.7685$ & $4.1910$ \\
\hline $d=6,500$ trees & $2.7713$ & $4.1932$ \\
\hline $d=7,500$ trees & $2.7712$ & $4.1923$ \\
\hline $d=8,500$ trees & $2.7712$ & $4.1924$ \\
\hline Canonical model, $S=1$ & $11.6717$ & $25.8051$ \\
\hline Canonical model, $S=2$ & $7.9865$ & $19.3870$ \\
\hline Canonical model, $S=3$ & $\mathbf{2.7442}$ & $\mathbf{4.1687}$ \\
\hline
\end{tabular}
\caption{Example results for Scenario \ref{scenario:mixture}}
    \label{tab:multipop}
\end{table}

Note that here the best \textit{multiresp} model outperforms the best XGB model. Note also that performance for the XGB model stops improving around depth 6 because only 6 of the 8 features behave differently between sub-responses.

\section{Failed approaches}

For completeness, we describe two experiments which did not demonstrate the tree depth - robustness connection.

\subsection{Highly nonlinear effects} 

We did not find a generator for which higher treedepth caused improved performance when modelling highly nonlinear effects from continuous features. As expected, high-depth models learned the `inner' parts of a w-shaped feature effect at the same rate as the `outer' parts, and lower-depth models fixated on the edges. But no scenario was found where this learning dynamic had a reliable impact on model performance.

\subsection{Outliers and mis-specified errors} 

The presence of outliers (i.e. the use of incorrect error distribution) appeared to lead to higher tree depth robustness, but the benefit was only consistent for some loss metrics, and only when tree depth was already $d>1$. That is: the MAE for $d=10$ would be consistently lower than for $d=5$ in the presence of sufficiently severe outliers, but we did not find a scenario where MAE was consistently lower than for $d=1$; moreover, extra complexity did not systematically improve RMSE at all.

(The ability of extra model complexity to lower MAE but not RMSE – by enabling it to ignore outliers – helps to explain the non-monotonic effect on RMSE in Table \ref{tab:multipop_plus_f}.)

%I mention these failures mostly in case they’re useful and to avoid publication bias, but also to show that there exist problems which do not increase optimal treedepth.